\PassOptionsToPackage{table}{xcolor}
\documentclass[journal]{IEEEtran}
\usepackage{changes}

\usepackage{graphicx}
\usepackage{float}
\usepackage{booktabs}
\usepackage{xcolor}
\usepackage{colortbl}
\usepackage{array}
\usepackage{stfloats}
\usepackage{amsmath}
\usepackage{verbatim}
\usepackage{enumitem}
\usepackage{multirow}
\usepackage{cuted}
\usepackage{algorithm}
\usepackage{algpseudocode}
\usepackage{amssymb}
\usepackage{url}
\usepackage{cite}

\usepackage{todonotes} % Load the todonotes package after setting the margin

\newcommand{\thistool}{PI-HNO}

\definecolor{coremetricbg}{gray}{0.93}
\newcolumntype{G}{>{\columncolor{coremetricbg}[\dimexpr\tabcolsep+1.4pt\relax][\dimexpr\tabcolsep+1.4pt\relax]}c}
\newcommand{\metricgroup}[2][0.86in]{%
  \begin{tabular}[c]{@{}c@{}}%
    \rule{0pt}{2.55ex}\textbf{#2}\\[-0.05ex]%
    \rule{#1}{0.35pt}%
  \end{tabular}%
}

\begin{document}

% \markboth{Preprint prepared for possible submission to IEEE Transactions on Power Electronics}%
% {Zhu \MakeLowercase{\textit{et al.}}: A Physics-Informed Hybrid Neural Operator for Transient Magnetization Prediction in High Frequency Power Magnetics}

\title{A Physics-Informed Hybrid Neural Operator for Transient Magnetization Prediction in High Frequency Power Magnetics}

\author{Yachao~Zhu,~\IEEEmembership{Student Member,~IEEE,}
        Qiujie~Huang,~\IEEEmembership{Student Member,~IEEE,}
        Sinan~Li,~\IEEEmembership{Member,~IEEE,}
        Yang~Li,~\IEEEmembership{Member,~IEEE,}
        Gang~Lei,~\IEEEmembership{Senior Member,~IEEE,}
        and~Jianguo~Zhu,~\IEEEmembership{Fellow,~IEEE}% <-this % stops a space
        
\thanks{This work is a preprint prepared for possible submission to IEEE Transactions on Power Electronics.}

\thanks{Yachao Zhu and Qiujie Huang are co-first authors. (Corresponding author: Qiujie Huang.)}
\thanks{Yachao Zhu and Gang Lei are with the School of Electrical and Data Engineering, the University of Technology Sydney, Ultimo, 2007, Australia (e-mail:Yachao.Zhu@uts.edu.au; Gang.Lei@uts.edu.au)}
\thanks{Qiujie Huang, Sinan Li, and Jianguo Zhu are with the School
of Electrical and Computer Engineering, the University of Sydney, Darlington,
NSW, 2008, Australia (e-mail:qiujie.huang@sydney.edu.au; sinan.li@sydney.edu.au; jianguo.zhu@sydney.edu.au)}
\thanks{Yang Li was with the National Railway Research and Design Institute of Signal and Communication, Beijing 100070, China (e-mail: 787511208@qq.com)}}% <-this % stops a space

% The paper headers
\markboth{PREPRINT PREPARED FOR POSSIBLE SUBMISSION TO IEEE TRANSACTIONS ON POWER ELECTRONICS}{}%

% make the title area
\maketitle

\begin{abstract}
Magnetic components in high-frequency, high-power-density converters are increasingly driven by non-sinusoidal flux-density waveforms with fast transitions, minor-loop operation, dc bias, and temperature variation. Under these conditions, steady-state core-loss formulas and single-valued material curves cannot fully capture transient magnetization responses. This work proposes the Physics-Informed Hybrid Neural Operator (\thistool), a compact material-specific neural model with $B$--$H$ energy consistency regularization for core-loss-oriented transient magnetization prediction. Given the measured $B(t)$--$H(t)$ history, the input $B(t)$ series over the prediction interval, and operating-condition information, \thistool\ predicts the $H(t)$ series and the corresponding reconstructed $B$--$H$ trajectory. \thistool\ integrates a local recurrent branch for boundary-state representation and rate-dependent response evolution with a Preisach-inspired global branch that extracts waveform-level hysteresis context. Evaluation on the MagNetX transient dataset using material-specific models for 14 ferrite materials demonstrates that \thistool\ balances sequence accuracy and $B(t)$--$H(t)$ energy consistency, achieving mean and 95th-percentile energy consistency errors of 1.92\% and 7.60\%, respectively, with only 4777 trainable parameters per model. Ablation study further demonstrates that the local, global, and energy-aware regularized components provide distinct contributions to transient magnetization prediction.
\end{abstract}


\IEEEpeerreviewmaketitle

\section{Introduction}

\IEEEPARstart{H}{igh-frequency} and high-power-density power converters enabled by wide-bandgap devices and advanced topologies impose increasing demands on magnetic components, including higher switching frequency, more complex excitation waveforms, and tighter thermal constraints \cite{ref_roshen,ref_how_magnet,ref_why_magnet}. Inductors and transformers directly affect converter volume, core loss, temperature rise, current stress, and efficiency, making accurate magnetic material modeling essential for compact power magnetic design \cite{ref_magnetx,ref_magnet_ai,teng2026overview}. However, modern converter excitations often involve fast transitions, dc bias, minor-loop operation, and temperature variation, where scalar core-loss models and single-valued material curves cannot capture the transient magnetic response required for time-domain analysis. For a given flux-density trajectory $B(t)$, the corresponding $H(t)$ response depends on magnetic history, reversal behavior, temperature, and rate-dependent effects, motivating compact models capable of predicting future $B$--$H$ trajectories from available history and excitation information.

These path-, rate-, and temperature-dependent behaviors have traditionally been described using empirical core-loss models and physics-based hysteresis formulations. Steinmetz-based models and their extensions provide efficient loss estimation for periodic waveforms, but cannot reconstruct transient $B$--$H$ trajectories under arbitrary excitation conditions \cite{ref_steinmetz,ref_mse,ref_gse,ref_igse}. Physics-based models, including Preisach, Jiles--Atherton, and generalized dynamic circuit models, provide improved interpretability by representing hysteresis memory and dynamic loss mechanisms \cite{ref_bertotti,ref_preisach,ref_jiles,ref_gdcm_i,ref_gdcm_ii}. However, these approaches often require material-specific parameter identification and may lose accuracy when operating frequency, temperature, bias, or geometry varies.

These challenges have motivated data-driven modeling of power magnetic materials. MagNet established a data foundation for $B$--$H$ loop reconstruction and core-loss modeling, while MagNet-AI introduced neural datasheet concepts for loop prediction, loss estimation, and material recommendation \cite{ref_how_magnet,ref_magnet_ai}. More recently, HARDCORE further explored time-domain $H(t)$ reconstruction and loss estimation from reconstructed trajectories \cite{ref_hardcore}. MagNetX extended transient magnetic modeling through time-domain excitation data and history-conditioned prediction \cite{ref_magnetx}.

In parallel, hybrid physics- and data-driven learning has been developed to improve physical consistency and reduce the ambiguity of purely data-driven magnetic models. Analytical core-loss-informed networks primarily focus on scalar loss prediction. In contrast, Preisach, Prandtl--Ishlinskii, and phenomenological-dynamics-based approaches encode path dependence and the multivalued nature of the $B$--$H$ characteristics. Neural operator approaches have also explored nonlinear mappings between excitation trajectories and magnetic responses, providing potential advantages for waveform-dependent modeling \cite{ref_empinn,ref_pinn_cross_attention,chandra2025generalizable,chandra2024magnetic,ref_mminn,ref_hdpi}. However, existing methods mainly focus on closed-loop loss regression, quasi-static hysteresis reconstruction, or constitutive mapping. A compact transient model that learns boundary-conditioned trajectory mappings while simultaneously providing accurate $H(t)$ prediction, $B$--$H$ energy consistency, and interpretable use of magnetization history and full-waveform $B(t)$ information remains insufficiently explored. This challenge is particularly important for finite prediction intervals, where reconstructed $B$--$H$ trajectories may be open and closed-cycle loss evaluation alone is insufficient.

Motivated by this gap, this work proposes the Physics-Informed Hybrid Neural Operator (\thistool), a compact neural material model for boundary-conditioned transient $B$--$H$ trajectory prediction. Unlike conventional sequence models that primarily learn pointwise temporal correlations, \thistool\ formulates transient magnetization prediction as a boundary-conditioned trajectory mapping between magnetic excitation, available history information, and future magnetic response. Given the measured magnetic history before the prediction interval, the input $B(t)$ trajectory over the interval, and operating-condition information, \thistool\ predicts the subsequent $H(t)$ response while improving the $B$--$H$ energy consistency of the reconstructed trajectory through physics-guided regularization.

The proposed physics-guided framework incorporates magnetic knowledge at three levels. First, the transient magnetic response is formulated based on a physics-motivated decomposition of history-dependent hysteresis behavior and rate-dependent dynamic effects, which guides the design of complementary learning branches. Second, magnetic memory and path dependence are represented using a Preisach-inspired global feature-extraction mechanism. Third, an energy-aware regularization term is introduced to improve the consistency of the reconstructed $B$--$H$ trajectory. Using these physics-guided representations and constraints, \thistool\ provides a compact framework for predicting the transient magnetic response. 

The main contributions are summarized as follows:
\begin{enumerate}

    \item A boundary-conditioned transient magnetization prediction framework is formulated for material-specific modeling, where the future $H(t)$ response is inferred from the measured magnetic history, the input $B(t)$ trajectory over the prediction interval, temperature, and temporal-position information.

    \item A compact hybrid neural architecture is developed by integrating local recurrent state propagation and global waveform-level context extraction for transient magnetic-response prediction. The proposed design enables simultaneous modeling of the evolution of the rate-dependent response and path-dependent hysteresis behavior under diverse excitation conditions.

    \item An energy-aware training objective is developed to improve the accumulated $B$--$H$ energy consistency of the predicted trajectory. Evaluations on the MagNetX transient dataset using 14 ferrite materials demonstrate that \thistool\ achieves a favorable trade-off among prediction accuracy, energy consistency, and model compactness.
\end{enumerate}

The remainder of this work is organized as follows. Section II defines the prediction problem and presents the physical background. Section III introduces the proposed \thistool\ and its energy-aware learning formulation. Section IV describes the experimental setup and evaluation metrics. Section V presents and discusses the comparative and ablation results. Conclusions are drawn in Section VI.

\section{Problem Formulation and Physical Background}

\subsection{Boundary-Conditioned Transient $H(t)$ Prediction}

\begin{figure}[!htbp]
    \centering
    \includegraphics[width=0.96\columnwidth]{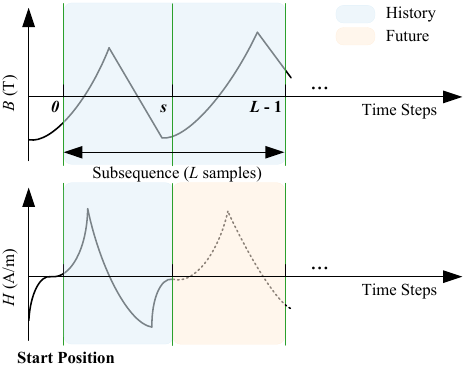}
    \caption{Input--output arrangement of the boundary-conditioned transient $H(t)$ prediction task.}
    \label{fig:input_output}
\end{figure}

The transient hysteresis prediction task is formulated as a boundary-conditioned trajectory mapping problem. As shown in Fig.~\ref{fig:input_output}, each transient $B$--$H$ sequence contains $L$ sampled time steps. A prediction boundary $s$ divides the sequence into a measured history interval, $0\leq t<s$, and a prediction interval, $s\leq t<L$. Before the boundary, both $B(t)$ and $H(t)$ are available and provide information about the magnetic state reached by the material. Over the prediction interval, the input $B(t)$ trajectory is known, whereas the corresponding $H(t)$ response is withheld from \thistool\ and used only as the training target or evaluation reference.

The objective is to learn a nonlinear trajectory mapping from the  magnetic history, excitation trajectory, and operating conditions to the future magnetization response:
\begin{equation}
    H_{\mathrm{pred}}(t)
    =
    \mathcal{G}_{\theta}
    \bigl(B(t), H_{\mathrm{hist}}(t), T\bigr),
    \qquad s\leq t<L ,
    \label{eq:prediction_mapping}
\end{equation}
where $\mathcal{G}_{\theta}$ denotes the learned nonlinear mapping from the input magnetic trajectories and operating conditions to the future magnetization response. $H_{\mathrm{hist}}(t)$ represents the measured magnetic history before the prediction boundary, and the complete $B(t)$ trajectory includes both the measured history and prediction intervals. Explicit frequency and waveform-periodicity labels are not provided as separate inputs.

The amount of measured history is controlled by the known-history ratio  as follows:
\begin{equation}
    \rho = \frac{s}{L},
    \label{eq:history_ratio}
\end{equation}
where a smaller $\rho$ indicates that less magnetization history is available. In contrast, a larger $\rho$ provides more available history and leads to a shorter prediction interval. The known-history ratio is varied across experiments to evaluate the proposed model under different levels of magnetic-history availability.

\subsection{Hysteresis Memory and Energy Consistency}

\begin{figure}[!t]
    \centering
    \includegraphics[width=\columnwidth]{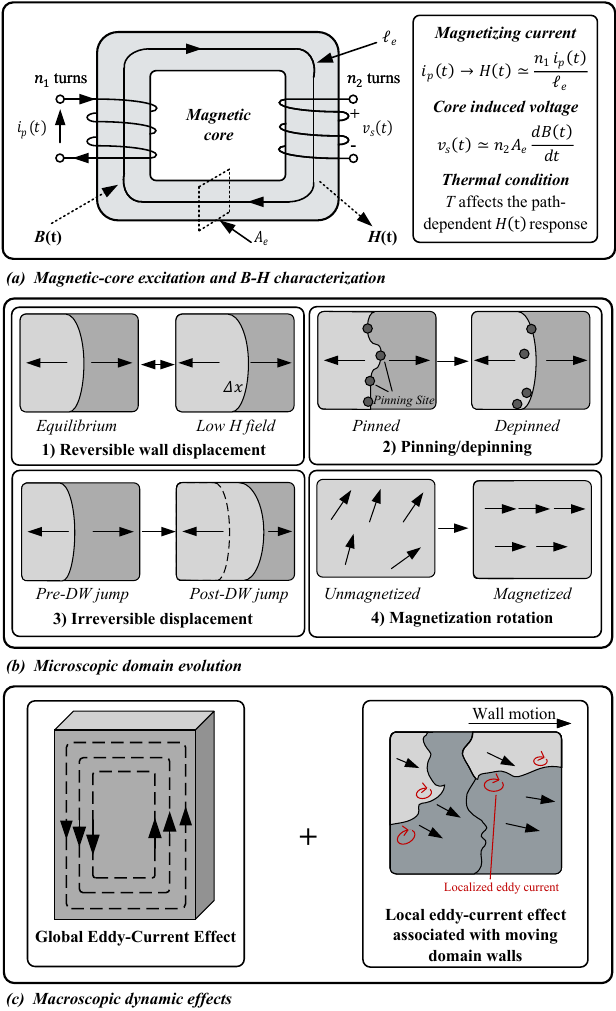}
    \caption{Physical interpretation of the magnetic response.}
    \label{fig:physical_background}
\end{figure}

The transient magnetic response considered in this work is governed by the excitation waveform, magnetization history, and rate-dependent effects. As shown in Fig.~\ref{fig:physical_background}, measured electrical quantities are converted into macroscopic $B(t)$--$H(t)$ trajectories that contain both hysteresis behavior and transient dynamic effects.

Following the two-winding characterization method in \cite{ref_magnetx}, $B(t)$ and $H(t)$ are obtained under the standard lumped-core approximation as
\begin{equation}
B(t)=B(t_0)+\frac{1}{n_2 A_e}\int_{t_0}^{t} v_s(\tau)\,\mathrm{d}\tau,
\qquad
H(t)\simeq \frac{n_1 i_p(t)}{\ell_e},
\label{eq:measured_bh_relations}
\end{equation}
where $v_s(t)$ and $i_p(t)$ denote the measured winding voltage and magnetizing current, respectively. $n_1$ and $n_2$ represent the primary and secondary windings. $A_e$ denotes effective cross-sectional area, and $\ell_e$ is magnetic flux path length. These relations link electrical measurements to macroscopic magnetic trajectories, but do not imply a unique instantaneous $B(t)$--$H(t)$ mapping.

At the material level, the magnetic flux density is related to the magnetic field and magnetization by
\begin{equation}
B(t)=\mu_0\bigl[H(t)+M(t)\bigr],
\label{eq:flux_field_magnetisation}
\end{equation}
where $M(t)$ represents the internal magnetization state. Unlike linear magnetic materials, the evolution of $M(t)$ in soft magnetic cores depends not only on the instantaneous flux density, but also on the previously experienced excitation trajectory. This memory effect arises from microscopic processes, including domain-wall displacement and magnetization rotation, that involve both reversible and irreversible mechanisms \cite{ref_jiles}.

Consequently, identical instantaneous operating points in the $B$--$H$ plane may lead to different subsequent magnetic responses when the underlying excitation histories differ. As illustrated in Fig.~\ref{fig:physical_background}(b), reversal points, flux-density extrema, dc bias, waveform asymmetry, and major- and minor-loop evolution can modify the internal magnetic state and in turn influence future magnetization trajectories. This path dependence explains why transient magnetization prediction requires not only the instantaneous excitation $B(t)$, but also sufficient information describing the previous magnetization history.

In addition to hysteresis memory, transient excitation introduces rate-dependent magnetic effects beyond quasi-static hysteresis. A time-varying flux density induces eddy currents within the magnetic core, generating electric fields that modify the measured magnetic field response. Moreover, delayed magnetic relaxation associated with domain-wall dynamics contributes to excess-loss-related dynamic behavior under high-frequency or rapidly varying excitation \cite{ref_bertotti}. Therefore, the transient field response can be conceptually interpreted as:
\begin{equation}
    H(t) = H_{\mathrm{hys}}\bigl(B(t),\xi(t),T\bigr) + H_{\mathrm{dyn}}(t),
    \label{eq:field_decomposition}
\end{equation}
where $\xi(t)$ denotes the internal magnetic state, $H_{\mathrm{hys}}$ represents the history-dependent hysteretic contribution, and $H_{\mathrm{dyn}}$ represents rate-dependent dynamic effects. In practical measurements, the individual contributions cannot be uniquely separated, because hysteresis evolution and dynamic effects are strongly coupled in the observed $B$--$H$ trajectory. Instead, this interpretation provides the physical motivation for incorporating both long-range magnetic history and local excitation variation into transient prediction models.

The predicted $H(t)$ response together with the input $B(t)$ trajectory defines a reconstructed $B$--$H$ path over the prediction interval. Therefore, evaluating transient magnetization prediction requires not only pointwise agreement in $H(t)$, but also the consistency of the accumulated $B$--$H$ work along the reconstructed trajectory. For a closed periodic hysteresis loop, the contour integral $\oint H\,\mathrm{d}B$ represents the dissipated core-loss energy density per cycle. In contrast, the boundary-conditioned prediction interval considered in this work generally forms an open trajectory $\mathcal{C}$,  as follows:
\begin{equation}
    W_{\mathcal{C}} = \int_{\mathcal{C}} H\,\mathrm{d}B .
    \label{eq:open_path_energy}
\end{equation}
For an open trajectory, $W_{\mathcal{C}}$ includes both recoverable magnetic work variation and dissipative contributions, and therefore it is not interpreted as a direct core-loss measurement. Instead, the difference between the predicted and measured values of $W_{\mathcal{C}}$ is used as an energy consistency measure of the reconstructed trajectory. This trajectory-level constraint complements the pointwise $H(t)$ prediction error and encourages physically consistent transient magnetization prediction.

\section{The Proposed Physics-Informed Hybrid Neural Operator}

This section presents \thistool\ for boundary-conditioned transient $B$--$H$ trajectory prediction. \thistool\ learns a boundary-conditioned trajectory operator that maps the measured magnetization history, the complete input $B(t)$ excitation trajectory, and temperature to the future $H(t)$ response. To achieve this trajectory-level prediction, \thistool\ combines a local recurrent branch for boundary-state representation and rate-dependent response evolution, a Preisach-inspired global branch for waveform-level hysteresis context extraction, and an energy-aware regularization term to improve the accumulated $B$--$H$ energy consistency of the reconstructed trajectory.

\subsection{Architecture Overview and Physics-Guided Decomposition}

\begin{figure*}[!t]
    \centering
    \includegraphics[width=\textwidth]{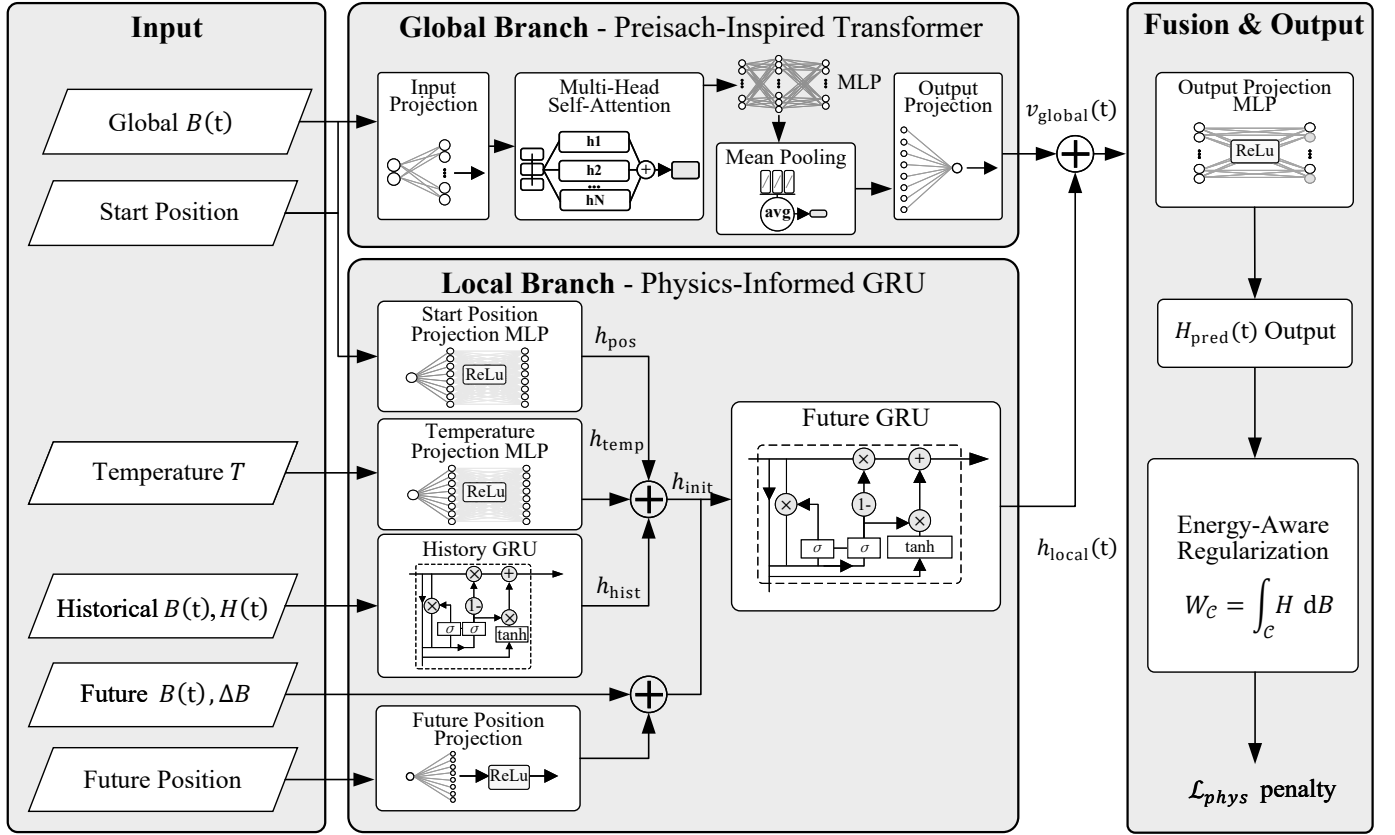}
    \caption{Architecture of the proposed \thistool.}
    \label{fig:architecture}
\end{figure*}

Fig.~\ref{fig:architecture} illustrates the overall architecture of \thistool. Motivated by the physical mechanisms discussed in Section~II, the proposed architecture incorporates a physics-guided decomposition of transient magnetic response into two complementary learning branches. Rather than explicitly identifying separate physical components, these branches are designed to learn representations related to history-dependent hysteresis behavior and rate-dependent magnetic dynamics.

\begin{itemize}
    \item The global branch is designed to extract waveform-level hysteresis context from the complete $B(t)$ trajectory. Inspired by the memory-weighting concept of the Preisach model, this branch learns path-dependent magnetic information associated with features such as ascending and descending flux-density trajectories, reversal points, and minor-loop evolution.
    
    \item The local recurrent branch is designed to learn the boundary-conditioned evolution of dynamic responses. It utilizes the measured $B(t)$--$H(t)$ history before the prediction interval and the step-to-step variation of the input $B(t)$ trajectory to propagate rate-dependent magnetic behavior across the prediction boundary.
\end{itemize}

Through this physics-guided architectural design, \thistool\ incorporates magnetic memory and dynamic-response information into the learning process rather than relying on a purely empirical input--output mapping. The representations learned from the local and global branches are jointly fused to predict the future $H_{\mathrm{pred}}(t)$ sequence. During training, an energy-aware regularization term is further introduced to improve the accumulated $B$--$H$ energy consistency of the predicted trajectory.

% ------------------
\subsection{Local Recurrent Branch for Dynamic Field Response}

The local branch is designed to learn the boundary-conditioned evolution of rate-dependent magnetic response under time-varying flux-density excitation. Following the physical interpretation discussed in Section~II-A, the transient magnetic response contains both history-dependent hysteresis behavior and rate-dependent dynamic effects. The local branch does not explicitly identify analytical eddy-current or excess-loss components. Instead, it learns an equivalent dynamic representation from measured magnetic trajectories. The observed $B(t)$--$H(t)$ history provides magnetic state information at the prediction boundary, while the input $B(t)$ waveforms over the prediction interval and their step-to-step variation determine the subsequent evolution of the excitation. A gated recurrent unit (GRU) encoder--decoder structure is adopted because its gated recurrence provides a compact, discrete-time representation of relaxation-type dynamics, in which the internal state is updated based on the current excitation while retaining relevant information from previous states.

\subsubsection{History GRU}

The history GRU reads the measured $B(t)$ and $H(t)$ sequences before the prediction boundary and compresses them into a hidden-state representation $h_{\mathrm{hist}}$. This representation provides initial magnetic-state information for future prediction, including the current field strength, the flux-density variation trend, remanent-state-related information, and reversal events in the measured history, thereby helping identify the active region of the hysteresis trajectory.

The use of a gated recurrent update is motivated by the relaxation-type behavior commonly observed in rate-dependent magnetization dynamics. An equivalent first-order relaxation model for an internal dynamic state can be expressed as:
\begin{equation}
\tau_t \frac{\mathrm{d}h(t)}{\mathrm{d}t}
=
h_{\mathrm{eq}}(x_t)-h(t),
\label{eq:relaxation_ode}
\end{equation}
where $h(t)$ represents an equivalent internal dynamic state, $h_{\mathrm{eq}}(x_t)$ is the excitation-dependent equilibrium state, $x_t$ denotes local excitation features, e.g., $B_t$, $\Delta B_t$, and $T$. $\tau_t$ represents an effective relaxation time constant. With a forward-Euler discretization, \eqref{eq:relaxation_ode} becomes:
\begin{equation}
h_t =
(1-\eta_t)h_{t-1}
+
\eta_t h_{\mathrm{eq}}(x_t),
\qquad
\eta_t=\frac{\Delta t}{\tau_t}.
\label{eq:euler_relaxation}
\end{equation}
This update shares the same weighted form as the GRU state equation,
\begin{equation}
h_t = (1-z_t)\odot h_{t-1} + z_t\odot \tilde{h}_t,
\label{eq:gru_update}
\end{equation}
where the update gate $z_t$ controls the contribution of previous states and current excitation information, and $\tilde{h}_t$ denotes the candidate hidden state determined by the current input. This analogy provides a physics-guided interpretation of the GRU update mechanism as a flexible discrete-time representation for rate-dependent magnetic response, rather than a direct analytical model of individual physical mechanisms such as eddy-current diffusion or domain-wall motion.

\subsubsection{Initial Hidden State Conditioning}

The initial state for future prediction cannot be determined solely from instantaneous magnetic variables because similar local magnetization states may exhibit different subsequent responses under different operating conditions. In particular, temperature variations can influence permeability, coercivity, conductivity, and relaxation behavior. In addition, the start-position feature identifies the location of the prediction boundary in the input waveform, allowing \thistool\ to distinguish repeated local excitation patterns that occur at different stages of the magnetization trajectory.

Therefore, the temperature $T$ and start-position feature are projected through two multilayer perceptrons (MLPs) and added to the hidden-state representation obtained from the history GRU:
\begin{equation}
\label{eq:hidden_init}
h_{\mathrm{init}} = h_{\mathrm{hist}} + h_{\mathrm{temp}} + h_{\mathrm{pos}},
\end{equation}
where $h_{\mathrm{temp}}$ and $h_{\mathrm{pos}}$ denote the learned projections of temperature and start position, respectively. The resulting state $h_{\mathrm{init}}$ initializes the future GRU, enabling the predicted response to incorporate temperature-dependent material behavior while using the start-position feature as contextual information for the prediction boundary.

\subsubsection{Future GRU}

The future GRU uses the same gated recurrent structure as the history GRU to propagate the learned dynamic representation over the prediction horizon. It is initialized by $h_{\mathrm{init}}$, which provides the boundary magnetic state inferred from the measured $B(t)$--$H(t)$ history. Beyond this boundary, the future response is driven by the input $B(t)$ trajectory. Therefore, at each prediction step, the future GRU receives the input flux density $B(t)$, its step-to-step increment $\Delta B(t)$, and a learned temporal-position embedding. The value of $B(t)$ specifies the instantaneous flux-density operating point, while $\Delta B(t)$ provides information about the local direction and rate of flux-density variation.

Including both inputs allows the future GRU to account for the fact that rate-dependent magnetic behavior is influenced not only by the instantaneous flux-density level, but also by how rapidly the excitation changes and by the magnetic state accumulated from the previous excitation history. With these inputs, the future GRU sequentially updates the hidden state and produces the local representation $h_{\mathrm{local}}(t)$. This representation encodes the rate-dependent magnetic response conditioned on the measured boundary state, the future excitation trajectory, and its local variation, and is subsequently fused with the global hysteresis representation for future $H(t)$ prediction.

\subsection{Global Attention Branch for Quasi-Static Hysteresis Context}

The local recurrent branch propagates the rate-dependent dynamic response step by step, but the quasi-static hysteresis component requires a broader description of the excitation history. As discussed in Section~II-B, hysteresis loss is path-dependent. It is affected by reversal points, flux-density extrema, dc flux-density offset, waveform asymmetry, and major- or minor-loop operation. These long-range waveform features can lead to different $H(t)$ responses, even when the local values of $B(t)$ and $\Delta B(t)$ are similar.

A natural way to represent path-dependent hysteresis memory is the Preisach model, which describes the quasi-static hysteretic response as a weighted superposition of elementary hysteresis units:
\begin{equation}
H_{\mathrm{hys}}(t)
=
\iint_{\alpha \ge \beta}
\mu(\alpha,\beta)\,
\gamma_{\alpha\beta}[B](t)
\,\mathrm{d}\alpha\,\mathrm{d}\beta ,
\label{eq:preisach_memory}
\end{equation}
where $\gamma_{\alpha\beta}[\cdot]$ denotes an elementary hysteresis operator with switching thresholds $\alpha$ and $\beta$, and $\mu(\alpha,\beta)$ represents the Preisach density function. This formulation describes hysteresis memory through the weighted aggregation of hysteron states over the Preisach plane.

\begin{figure}[!t]
    \centering
    \includegraphics[width=1\columnwidth]{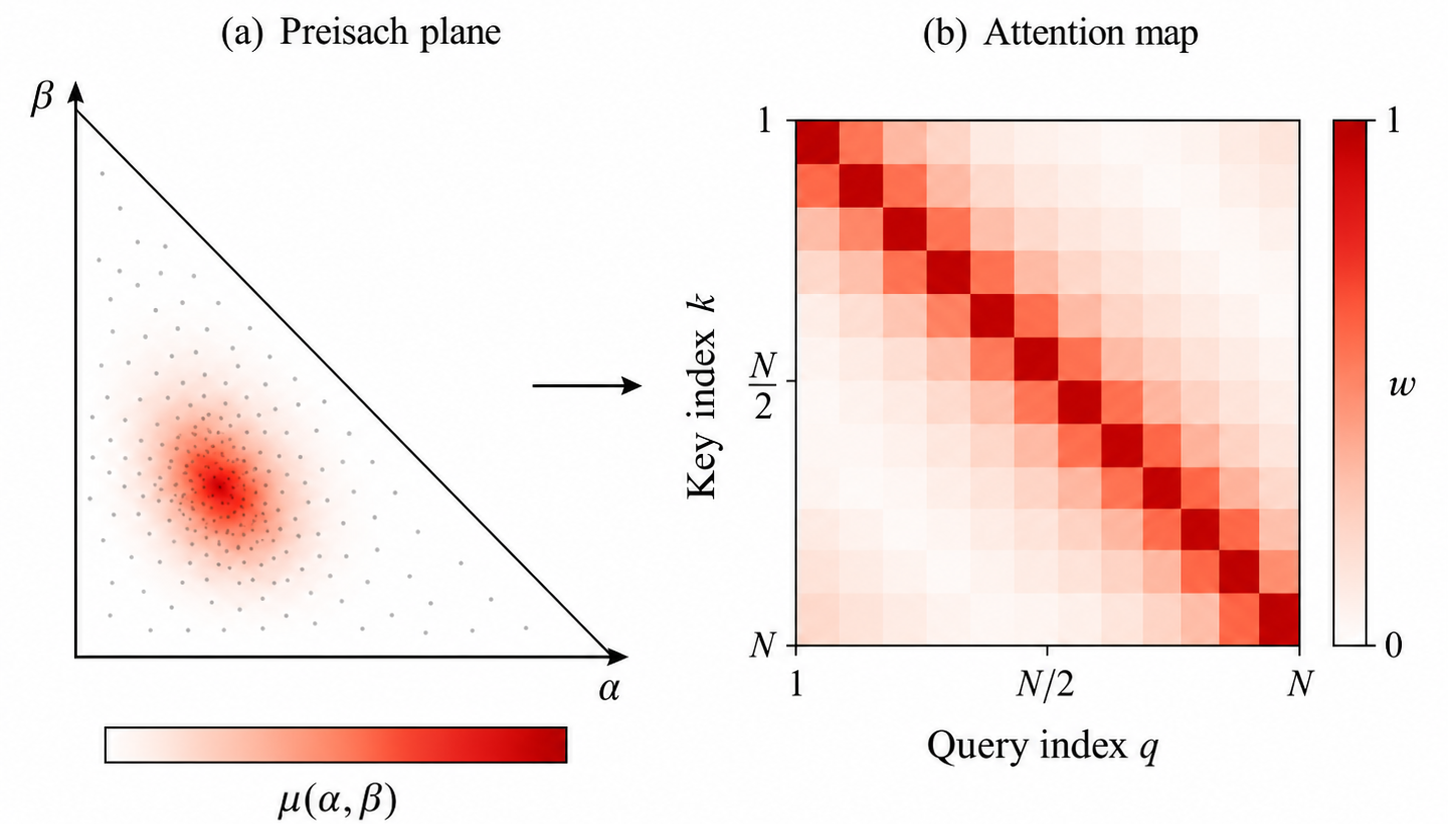}
    \caption{Preisach-inspired global attention mechanism for extracting hysteresis-related waveform context.}
    \label{fig:preisach_attention}
\end{figure}

Motivated by the memory aggregation concept underlying Preisach-type hysteresis descriptions, \thistool\ introduces a Transformer-based global branch, as illustrated in Fig.~\ref{fig:preisach_attention}. The branch does not explicitly identify the Preisach density or implement physical hysterons. Instead, it employs multi-head self-attention as a data-driven memory aggregation mechanism, in which the learned attention weights capture the relevance of different portions of the excitation trajectory to predicting the transient magnetic response. For a projected flux-density sequence, the attention weight from the query position $q$ to a key position $k$ is
\begin{equation}
A_{qk}
=
\frac{
\exp \left( Q_q K_k^{\mathrm{T}}/\sqrt{d_k} \right)
}{
\sum_{\ell=1}^{N}
\exp \left( Q_q K_{\ell}^{\mathrm{T}}/\sqrt{d_k} \right)
},
\label{eq:attention_weight}
\end{equation}
and the corresponding attention output $z_q$ is
\begin{equation}
z_q = \sum_{k=1}^{N} A_{qk} V_k .
\label{eq:attention_output}
\end{equation}
where, $q$ denotes the query position at which the attention output is computed, and $k$ denotes the key-value position being attended to. $Q_q$, $K_k$, and $V_k$ are the query, key, and value vectors obtained through learned linear projections of the input embedding, respectively. $d_k$ is the dimension of the key vector, which is used to scale the dot-product similarity. $A_{qk}$ is the normalized attention weight that measures the contribution of position $k$ to position $q$. Therefore, the attention output can be regarded as a data-driven analog of Preisach-type weighted superposition, where learned weights over sampled positions of the input $B(t)$ waveform provide a discrete representation of the path-dependent magnetization context.

Following this interpretation, the global branch extracts the path-dependent magnetization context from the complete input $B(t)$ trajectory. The $B(t)$ sequence and the broadcast start-position feature are projected and passed through the self-attention block, where different portions of the excitation waveform are compared directly. Mean pooling over the resulting time-step representations produces the global vector $v_{\mathrm{global}}(t)$, which summarizes branch information, reversal locations, flux-density extrema, dc offset, and minor-loop structure. This quasi-static context complements the local dynamic branch by distinguishing prediction segments with similar instantaneous $B(t)$ and $\Delta B(t)$ but different magnetization histories.

\subsection{Branch Fusion and Output Layer}

The local and global branches provide complementary representations for future $H(t)$ prediction. The local branch produces a time-dependent representation $h_{\mathrm{local}}(t)$ through sequential GRU updates, reflecting rate-dependent dynamic field behavior. In contrast, the global branch produces $v_{\mathrm{global}}(t)$ from the complete input $B(t)$ trajectory, providing sequence-level context for quasi-static, path-dependent magnetization. As shown in Fig.~\ref{fig:architecture}, these two representations are combined by additive fusion and mapped to $H_{\mathrm{pred}}(t)$ through output projection:
\begin{equation}
u(t) = h_{\mathrm{local}}(t) + v_{\mathrm{global}}(t),
\label{eq:additive_fusion}
\end{equation}
\begin{equation}
H_{\mathrm{pred}}(t) =
\mathrm{MLP}_{\mathrm{out}}\!\left(u(t)\right).
\label{eq:fusion_output}
\end{equation}

This additive fusion is motivated by the loss-separation picture discussed in Section~II-B, in which the measured field response can be interpreted as the combined effect of quasi-static hysteresis and dynamic eddy-current components. In \thistool, these components are not explicitly identified as analytical field terms. Instead, the local and global branches learn equivalent physics-guided representations that are summed in the latent space and mapped to the predicted future $H_{\mathrm{pred}}(t)$ sequence.

After the output projection, $H_{\mathrm{pred}}(t)$ and the input $B(t)$ trajectory form the predicted $B$--$H$ trajectory over the prediction interval. During training, the energy-aware regularization term shown in Fig.~\ref{fig:architecture} penalizes inconsistency in the accumulated $B$--$H$ energy of this trajectory. For inference, \thistool\ directly outputs $H_{\mathrm{pred}}(t)$ from the measured history, input $B(t)$ trajectory, temperature, and temporal-position features.

\subsection{Energy-Aware Training Objective}\label{sec:loss}

\thistool\ is trained by combining pointwise waveform matching with a soft $B$--$H$ energy consistency regularization term:
\begin{equation}\label{eq:loss_total}
    \mathcal{L}
    =
    \mathcal{L}_{\mathrm{norm}}
    +
    \lambda_{\mathrm{phys}}\mathcal{L}_{\mathrm{phys}},
\end{equation}
where $\mathcal{L}_{\mathrm{norm}}$ is the smooth-$L_1$ loss between $H_{\mathrm{pred}}(t)$ and the measured reference $H_{\mathrm{meas}}(t)$ in normalized units, and $\lambda_{\mathrm{phys}}$ weights the $B$--$H$ energy consistency term.

The energy consistency term is computed after converting the normalized $B$ and $H$ values back to physical units. With the input $B(t)$ trajectory over the prediction interval, the path integral of $H\,\mathrm{d}B$ is approximated by the trapezoidal rule for both the predicted and measured $B$--$H$ trajectories. The resulting open-path quantities, denoted as $W_{\mathrm{pred}}$ and $W_{\mathrm{meas}}$, are used to regularize $B$--$H$ trajectory consistency rather than to directly evaluate closed-cycle core loss. Since the prediction interval generally corresponds to an open $B$--$H$ path rather than a closed hysteresis loop, this energy consistency is imposed only as a soft regularization:
\begin{equation}\label{eq:loss_phys}
    \mathcal{L}_{\mathrm{phys}}
    =
    \frac{\bigl|W_{\mathrm{pred}}-W_{\mathrm{meas}}\bigr|}
    {\max\bigl(|W_{\mathrm{meas}}|,\epsilon\bigr)}
    +
    \gamma\max\bigl(0,-W_{\mathrm{pred}}\bigr),
\end{equation}
where the first term penalizes inconsistency between the predicted and measured accumulated $B$--$H$ energy, the second term introduces a one-sided penalty for negative accumulated energy, $\epsilon$ avoids numerical instability when the reference energy approaches zero, and $\gamma$ controls the strength of this penalty. This composite regularization preserves supervision from the measured $H(t)$ sequence while encouraging physically consistent reconstructed $B$--$H$ trajectories.

% Therefore, the proposed energy term should be interpreted as a trajectory-level physical regularization rather than a direct loss-prediction objective. It constrains the predicted magnetization path to preserve the energy-related characteristics of the measured response.

%%*************************************************************************
\section{Experimental Setup and Evaluation Metrics}

\subsection{Dataset and Prediction-Sequence Construction}

The experiments use transient magnetic data from MagNetX \cite{ref_how_magnet,ref_magnetx}, following the history-conditioned prediction formulation defined in Section~II-A. MagNetX measures $B(t)$ and $H(t)$ using a two-winding magnetic characterization setup, where $B(t)$ is obtained from secondary-side voltage integration and $H(t)$ is obtained from primary-side current measurement. The original transient measurements are collected under randomly varying duty-cycle and frequency-transition excitations. The sampling window is adjusted based on the excitation frequency to capture multiple waveform cycles, with each measurement containing 100,000 samples and a 20-MHz bandwidth limit. Subsequently, each measurement is down-sampled to 10,000 samples by averaging every ten consecutive samples \cite{ref_magnetx}.

The evaluated material set contains 14 ferrite materials: 3C90, 3C92, 3C94, 3C95, 3E6, 3F4, 77, 78, N27, N30, N49, N87, FEC014, and T37. Most materials include excitation frequencies of 50, 80, 125, 200, 320, 500, and 800~kHz, except 50 and 80~kHz are not available for N49. The measured temperatures are 25, 50, and 70~$^{\circ}$C.

For the experiments in this work, the down-sampled transient waveforms are further organized into fixed-length sequences with $L=1000$ time steps, consistent with the notation in \eqref{eq:history_ratio}. For each sequence, the prediction boundary $s$ is selected according to the test known-history ratios $\rho\in\{0.1,0.5,0.9\}$. The same sequence construction is used for all evaluated models, while each model retains its released input and feature definitions.

\subsection{Training, Validation, and Testing Setup}

The training, validation, and testing datasets are constructed separately for each material from the processed MagNetX transient measurements. This material-specific formulation is adopted because different magnetic materials exhibit distinct hysteresis characteristics, dynamic magnetic behaviors, and temperature-dependent responses. Therefore, this work focuses on developing compact transient magnetization models tailored to individual materials, rather than a single universal model across heterogeneous magnetic materials. Within each material, the proposed model is evaluated under diverse excitation waveforms, frequency conditions, temperatures, and available magnetic-history lengths.

For each frequency-temperature group, the measured $B(t)$, $H(t)$, and temperature waveforms are first screened to remove invalid samples with inconsistent incremental magnetic responses. The remaining waveforms are then sampled with a fixed random seed and divided into nonoverlapping 1000-time-step sequences. The resulting material-wise datasets contain aligned $B$, $H$, and temperature arrays for supervised training.

For training, each 1000-time-step sequence is further divided into local prediction segments. Within each segment, the most recent $m=100$ samples before the prediction boundary are used as the measured $B(t)$--$H(t)$ history for the history GRU, while the subsequent samples provide the prediction target for the future GRU. The complete 1000-time-step $B(t)$ sequence is retained as the input to the global branch. For materials showing weaker validation performance at low peak-to-peak flux density ($B_{\mathrm{pp}}$), additional low-$B_{\mathrm{pp}}$ training sequences are included to improve coverage of small-signal transient behavior. This data augmentation is determined solely from the corresponding training and validation results and does not involve the test data.

A validation subset is held out from the training sequences and is used exclusively for model selection, early stopping, and hyperparameter tuning. The test sequences remain independent throughout training and validation and are used only for final performance evaluation. For the test known-history ratios $\rho\in\{0.1,0.5,0.9\}$, test sequences are constructed according to the corresponding prediction boundary $s=\rho L$. The same test set for each $\rho$ is applied to all evaluated models within each material, ensuring a consistent comparison under different available-history conditions.

All baseline models are trained and evaluated using the same material-specific data construction, training/validation/testing splits, input $B(t)$ trajectories, and evaluation metrics, ensuring that performance differences primarily reflect the model architecture rather than experimental settings while retaining model-specific training procedures. MagLearn2, HARDCORE, magnetization mechanism-inspired neural network (MMINN) and long short-term memory (LSTM) retain their released model architectures, inputs, and features.

\subsection{Implementation Details of \thistool}

The history GRU has a hidden width of 8 and one layer, and the future GRU uses the same hidden width and depth. The global branch is a single-layer Transformer with width 8, two attention heads, and feed-forward width 64. The temperature and start-position projection MLPs use sizes 1--8--8, the future-position projection uses sizes 1--8, and the output projection MLP uses sizes 8--128--1 with a dropout rate of 0.1.

For training, each 1000-time-step training sequence is divided into five consecutive nonoverlapping 200-time-step local training segments. In each segment, the first 100 time steps provide the measured history to the history GRU, and the last 100 time steps form the prediction target for the future GRU. The complete 1000-time-step $B(t)$ sequence serves as the global branch input for all five local segments. The start-position feature is the segment starting index normalized by the sequence length, and the future-position features are the normalized indices of the 100 target time steps.

Training minimizes the composite objective in Section~\ref{sec:loss}, with $\lambda_{\mathrm{phys}}=0.1$, $\gamma=0.01$, and $\epsilon=10^{-6}$. A batch size of 64 is used for at most 1000 epochs with the AdamW optimizer and a one-cycle learning-rate schedule with a peak learning rate of $10^{-2}$. The weight decay is $5\times10^{-4}$, the gradient norm is clipped at 0.5, and an exponential moving average with decay 0.999 is used for validation. Training stops when the validation loss does not improve for 80 consecutive epochs.

\subsection{Baseline Models and Ablation Settings}
\label{sec:comparison_ablation}

To evaluate the proposed \thistool, six baseline models are implemented under the same data construction and evaluation procedure. These models are selected to cover major categories of transient magnetic modeling approaches, including sequence-to-sequence recurrent models, convolutional neural models, physics-inspired neural architectures, and attention-based sequence models. Existing neural operator approaches have primarily been developed to learn nonlinear mappings between input and output functions, with an emphasis on general operator approximation, discretization-independent learning, or continuous field modeling. In contrast, this work focuses on compact, boundary-conditioned prediction of transient magnetization trajectories under material-specific operating conditions, with computational efficiency and trajectory reconstruction accuracy as the primary objectives. Therefore, task-relevant magnetic response prediction models are selected as the primary references to provide meaningful, computationally comparable baselines.

The baseline models include the sequence-to-sequence MagLearn2 model \cite{ref_maglearn2}, an LSTM model based on the MagNet Challenge 2 demonstration setting \cite{ref_magnetx}, the residual dilated-convolution HARDCORE model for time-domain $H(t)$ and loss estimation \cite{ref_hardcore}, the magnetization-mechanism-inspired neural network MMINN \cite{ref_mminn}, a GRU model, and a Transformer model. The GRU model is configured with a parameter size close to that of \thistool\ to evaluate the contribution of the proposed hybrid architecture beyond parameter size. In contrast, the Transformer model uses the same Transformer width, head count, and depth as the global branch of \thistool\ with an independent output head to evaluate the contribution of the local recurrent branch. HARDCORE and MMINN retain their released architectures, with their data interfaces adapted to the MagNetX dataset. The scalar loss-correction branch is not instantiated for HARDCORE, as the MagNetX transient dataset does not provide an equivalent per-profile volumetric power-loss target. The model configurations are summarized in Table~\ref{tab:reference_model_configuration}.

\begin{table}[!t]
\centering
\caption{Configurations of \thistool\ and the baseline models}
\label{tab:reference_model_configuration}
\footnotesize
\setlength{\tabcolsep}{2.4pt}
\renewcommand{\arraystretch}{1.08}
\begin{tabular}{@{}ll@{}}
\toprule
\textbf{Model} & \textbf{Structure} \\
\midrule
\thistool & $2\mathrm{GRU}(8)\!\times\!1+\mathrm{Transformer}(8,2)\!\times\!1+\mathrm{MLP}(128)$ \\
MagLearn2 & $2\mathrm{LSTM}(128)\!\times\!3+\mathrm{MLP}(256,128)$ \\
LSTM & $2\mathrm{LSTM}(12)\!\times\!1+\mathrm{MLP}(26,8,4)$ \\
HARDCORE & $\mathrm{TCN}(5,12,8,1;k{=}9)+\mathrm{MLP}(11)$ \\
MMINN & $\mathrm{RNN}(30)+\mathrm{MLP}(30)$ \\
GRU & $2\mathrm{GRU}(15)\!\times\!1+\mathrm{MLP}(128)$ \\
Transformer & $\mathrm{Transformer}(8,2)\!\times\!1+\mathrm{MLP}(64)$ \\

% \thistool & $\mathrm{GRU}(8)\!\times\!2+\mathrm{Transformer}(8,2)\!\times\!1+\mathrm{MLP}(128)$ \\
% MagLearn2 & $2\mathrm{LSTM}(128)\!\times\!3+\mathrm{MLP}(256,128)$ \\
% LSTM & $2\mathrm{LSTM}(12)\!\times\!1+\mathrm{MLP}(26,8,4)$ \\
% HARDCORE & $\mathrm{TCN}(5,12,8,1;k{=}9)+\mathrm{MLP}(11,11)$ \\
% MMINN & $\mathrm{RNN}(30)+\mathrm{MLP}(30)$ \\
% GRU & $\mathrm{GRU}(15)\!\times\!2+\mathrm{MLP}(128)$ \\
% TF & $\mathrm{TF}(8,2)\!\times\!1+\mathrm{MLP}(64)$ \\
\bottomrule
\end{tabular}
\par\vspace{0.6mm}
\end{table}

\begin{figure}[!t]
    \centering
    \includegraphics[width=0.6\columnwidth]{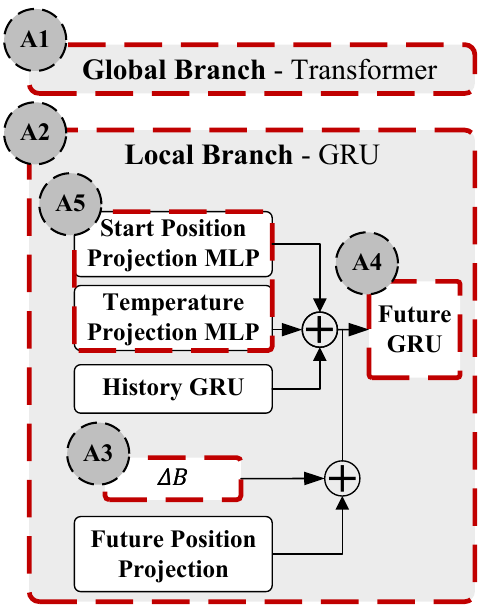}
    \caption{Visualization of ablation variants A1--A5 of \thistool.}
    \label{fig:ablation_design}
\end{figure}

The ablation study is designed to identify the contribution of each major architectural or conditioning component in \thistool. All ablation variants use the same dataset construction, training setup, and evaluation metrics as the full model, except for the stated modification. As shown in Fig.~\ref{fig:ablation_design}, five variants are evaluated. A1 removes the global branch to test the contribution of sequence-level hysteresis context beyond the local measured history. A2 removes the local recurrent branch to test the role of local dynamic-state propagation and boundary-state information. A3 removes the incremental $\Delta B$ input to assess the importance of local flux-density variation for rate-dependent field prediction. A4 removes the future GRU to test whether the magnetic state must be propagated sequentially along the input $B(t)$ trajectory. A5 changes how temperature and start-position information are used by feeding these scalar conditions directly to the future GRU rather than using them to initialize the hidden state, thereby conditioning the location rather than removing the conditions entirely.

\subsection{Evaluation Metrics}
\label{sec:evaluation_metrics}

Four metrics are used to evaluate complementary aspects of the predicted transient magnetization response. The normalized sequence error $E_{\mathrm{seq}}$ measures the overall waveform agreement between the predicted and measured $H(t)$ sequences. The peak-field error, $E_{H_{\mathrm{pk}}}$, evaluates whether the model correctly predicts the maximum field magnitude, which is relevant to the magnetic saturation margin and current-stress estimation. The relative $B$--$H$ energy consistency error $E_{\mathrm{ene}}$ measures the consistency of the accumulated $B$--$H$ work along the predicted trajectory. This quantity corresponds to core-loss evaluation only when the reconstructed trajectory forms a closed cycle. For the open trajectories considered in this work, it is used as a trajectory-level energy consistency measure. The energy-sign mismatch rate $\eta_{\mathrm{sgn}}$ counts the fraction of test sequences for which the predicted and measured accumulated $B$--$H$ work have opposite signs, indicating an inconsistent direction of the accumulated work over the prediction interval. As defined in Section~II-A, $s$ is the prediction boundary determined by the known-history ratio in \eqref{eq:history_ratio}. All metrics are computed over the prediction interval $k=s,\ldots,L-1$. For clarity, the sequence index is omitted in the following definitions. For a given test sequence, $H_{\mathrm{meas},k}$, $H_{\mathrm{pred},k}$, and $B_k$ denote the measured field strength, predicted field strength, and input flux density at time step $k$, respectively.

The normalized sequence error is
\begin{equation}
    E_{\mathrm{seq}}
    =
    \frac{
    \sqrt{\dfrac{1}{L-s}
    \sum_{k=s}^{L-1}
    \left(H_{\mathrm{pred},k}-H_{\mathrm{meas},k}\right)^2}
    }{
    \max\!\left(
    \sqrt{\dfrac{1}{L-s}
    \sum_{k=s}^{L-1}H_{\mathrm{meas},k}^{2}},
    \epsilon\right)
    }\times100\% .
    \label{eq:seq_metric}
\end{equation}
This metric measures the pointwise waveform error over the prediction horizon. The peak-field error is
\begin{equation}
    E_{H_{\mathrm{pk}}}
    =
    \frac{
    \left|
    \max\limits_{s\leq k\leq L-1}\left|H_{\mathrm{pred},k}\right|
    -
    \max\limits_{s\leq k\leq L-1}\left|H_{\mathrm{meas},k}\right|
    \right|
    }{
    \max\!\left(
    \max\limits_{s\leq k\leq L-1}\left|H_{\mathrm{meas},k}\right|,
    \epsilon\right)
    }\times100\% .
    \label{eq:hpk_metric}
\end{equation}
This metric evaluates the predicted maximum field magnitude.

For implementation, the open-path quantity in \eqref{eq:open_path_energy} is evaluated over the prediction interval using the trapezoidal rule. For the predicted trajectory,
\begin{equation}
    W_{\mathrm{pred}}
    =
    \sum_{k=s}^{L-2}
    \frac{H_{\mathrm{pred},k+1}+H_{\mathrm{pred},k}}{2}
    \left(B_{k+1}-B_k\right).
    \label{eq:wpred_metric}
\end{equation}
The measured counterpart $W_{\mathrm{meas}}$ is obtained by replacing $H_{\mathrm{pred},k}$ with $H_{\mathrm{meas},k}$ in \eqref{eq:wpred_metric}. Since the prediction interval generally forms an open $B$--$H$ trajectory, these quantities are not interpreted as closed-cycle core-loss energy. Their discrepancy is instead used as an energy-related consistency measure that complements the pointwise sequence error in $H(t)$. The relative energy error is
\begin{equation}
    E_{\mathrm{ene}}
    =
    \frac{
    \left|W_{\mathrm{pred}}-W_{\mathrm{meas}}\right|
    }{
    \max\!\left(\mathcal{S}_{\mathrm{ene}},\varepsilon_{\mathrm{ene}}\right)
    }\times100\% ,
    \label{eq:energy_metric}
\end{equation}
where $\mathcal{S}_{\mathrm{ene}}$ is the absolute accumulated-energy scale of the test sequence and $\varepsilon_{\mathrm{ene}}$ is a small positive constant. This metric evaluates the consistency of the accumulated $B$--$H$ work along the predicted trajectory.

For an evaluation pool containing $N_{\mathrm{seq}}$ test sequences, the energy-sign mismatch rate is
\begin{equation}
    \eta_{\mathrm{sgn}}
    =
    \frac{1}{N_{\mathrm{seq}}}
    \sum_{j=1}^{N_{\mathrm{seq}}}
    \mathbf{1}\!\left[
    W_{\mathrm{pred}}^{(j)}W_{\mathrm{meas}}^{(j)}<0
    \right]\times100\% ,
    \label{eq:sign_metric}
\end{equation}
where $\mathbf{1}[\cdot]$ is the indicator function. This diagnostic identifies cases where the predicted accumulated work has the opposite sign to the measured reference.

For each material, all valid test-sequence evaluations are pooled across the evaluated known-history ratios, frequency levels, and temperatures. The distributions of $E_{\mathrm{seq}}$ and $E_{\mathrm{ene}}$ are summarized by their mean and 95th-percentile values, $E_{H_{\mathrm{pk}}}$ is reported by its mean value, and $\eta_{\mathrm{sgn}}$ is reported as the proportion of sign-mismatched test sequences. Overall results are obtained by taking the unweighted arithmetic mean of the corresponding material-wise statistics across the 14 materials.

%%*************************************************************************

\section{Results and Discussion}

This section presents the experimental results for \thistool\ and discusses its prediction accuracy, energy consistency, comparisons with baseline models, ablation results, and model compactness. All results are obtained using the dataset construction, training setup, and evaluation metrics described in Section~IV.

\subsection{Overall Prediction Performance}

Table~\ref{tab:baseline_material} summarizes the material-wise prediction performance of \thistool. The overall statistics are computed as unweighted averages of the 14 material-wise results. Across all evaluated materials, \thistool\ achieves mean sequence and energy errors of 13.40\% and 1.92\%, respectively, with corresponding 95th-percentile values of 31.39\% and 7.60\%. The mean peak-field error is 6.12\%, and the energy-sign mismatch rate is 1.91\%, indicating that most predicted trajectories preserve the direction of accumulated $B$--$H$ work over the prediction interval.

\begin{table}[!t]
\centering
\caption{Material-wise and overall sequence, energy, peak-field errors and energy-sign mismatch of \thistool}
\label{tab:baseline_material}
\footnotesize
\setlength{\tabcolsep}{3.4pt}
\renewcommand{\arraystretch}{1.06}
\begingroup
\setlength{\aboverulesep}{0pt}
\setlength{\belowrulesep}{0pt}
\begin{tabular}{@{}lGGGGcc@{}}
\toprule
\multirow{2}{*}{\textbf{Material}}
& \multicolumn{2}{G}{\metricgroup{Sequence Error}}
& \multicolumn{2}{G}{\metricgroup{Energy Error}}
& \begin{tabular}[c]{@{}c@{}}\textbf{Peak}\\\textbf{Field}\\\textbf{Error}\end{tabular}
& \begin{tabular}[c]{@{}c@{}}\textbf{Energy}\\\textbf{Sign}\\\textbf{Mismatch}\end{tabular} \\
& Mean (\%) & 95th (\%) & Mean (\%) & 95th (\%) & (\%) & (\%) \\
\midrule
3C90 & 10.72 & 27.84 & 1.66 & 6.67 & 8.58 & 1.47 \\
3C92 & 18.46 & 38.15 & 1.07 & 3.93 & 7.83 & 2.42 \\
3C94 & 13.51 & 39.07 & 1.90 & 7.44 & 4.76 & 1.19 \\
3C95 & 7.31 & 16.56 & 1.13 & 4.54 & 3.86 & 0.90 \\
3E6 & 17.37 & 36.45 & 1.27 & 5.40 & 6.97 & 1.63 \\
3F4 & 9.84 & 24.39 & 3.01 & 9.40 & 3.10 & 1.63 \\
77 & 14.01 & 31.79 & 1.51 & 5.46 & 8.24 & 2.25 \\
78 & 13.54 & 30.32 & 2.49 & 9.75 & 5.69 & 1.59 \\
N27 & 14.94 & 32.35 & 1.76 & 7.06 & 8.84 & 3.25 \\
N30 & 20.41 & 47.97 & 1.86 & 7.92 & 6.74 & 2.16 \\
N49 & 12.61 & 27.70 & 4.08 & 16.36 & 6.61 & 4.13 \\
N87 & 17.45 & 44.24 & 2.07 & 7.88 & 5.96 & 1.70 \\
FEC014 & 8.33 & 22.63 & 1.92 & 9.12 & 3.52 & 1.38 \\
T37 & 9.10 & 19.97 & 1.17 & 5.48 & 4.95 & 1.08 \\
\midrule
Average & 13.40 & 31.39 & 1.92 & 7.60 & 6.12 & 1.91 \\
\bottomrule
\end{tabular}
\endgroup
\end{table}

The material-wise results show that the difficulty of sequence prediction varies noticeably across materials. The lowest mean sequence errors are obtained for 3C95, FEC014, T37, and 3F4, all below 10\%, whereas N30, 3C92, N87, and 3E6 show larger mean sequence errors above 17\%. The 95th-percentile sequence error follows a similar trend, with N30 and N87 yielding the largest values. In contrast, the energy error remains low for most materials: the mean energy error is below 2.5\% for 12 of the 14 materials, and below 1.3\% for 3C92, 3C95, 3E6, and T37. The main exception is N49, which has the highest mean and 95th-percentile energy errors of 4.08\% and 16.36\%, respectively, and also the highest energy-sign mismatch rate of 4.13\%. This is likely related to its reduced frequency coverage, since the 50 and 80~kHz conditions are not available for N49.

\begin{figure*}[!t]
    \centering
    \includegraphics[width=\textwidth]{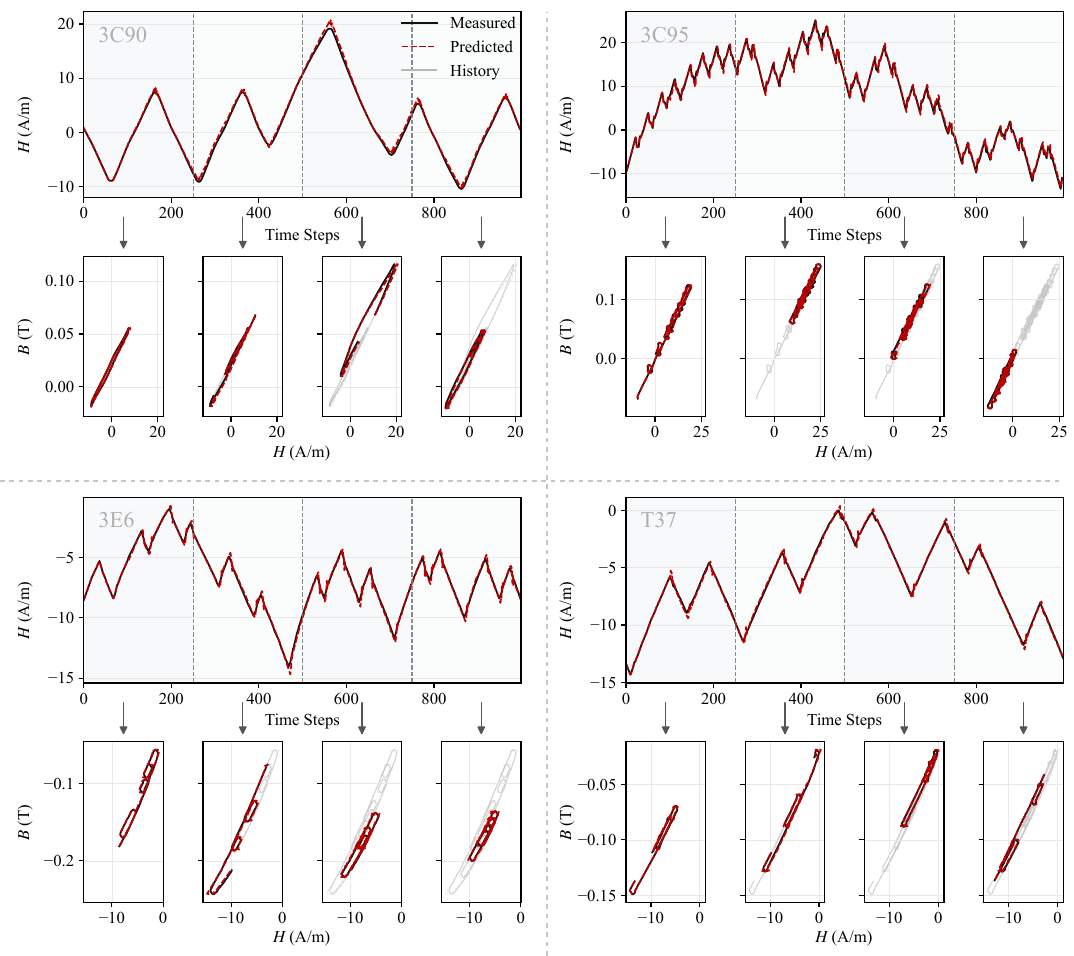}
    \caption{Representative future $H(t)$ prediction results for 3C90, 3C95, 3E6, and T37.}
    \label{fig:representative_prediction}
\end{figure*}

Fig.~\ref{fig:representative_prediction} shows representative predictions for 3C90, 3C95, 3E6, and T37. The upper plots compare the measured, predicted, and historical $H(t)$ sequences, while the lower plots show selected predicted and measured $B$--$H$ segments. The predictions follow the main waveform trends and turning points, and the corresponding $B$--$H$ segments preserve the measured loop direction and branch location. These examples visually support the sequence and energy-error statistics in Table~\ref{tab:baseline_material}.

The difference between sequence-error and energy-error trends shows that waveform fitting and accumulated-energy consistency are complementary criteria that ensure the measured $B$--$H$ path is well followed to reproduce the accumulated energy, even when the pointwise sequence error varies across materials.

\subsection{Comparison With Baseline Models}

Table~\ref{tab:comparison_models} compares \thistool\ with six models. MagLearn2 achieves the lowest overall sequence errors, with mean and 95th-percentile values of 12.88\% and 28.47\%, respectively. The corresponding values of \thistool\ are 13.40\% and 31.39\%, indicating that \thistool\ does not achieve the best performance in terms of pointwise waveform fitting alone. However, transient magnetization modeling requires not only accurate $H(t)$ sequence prediction, but also a physically consistent reconstruction of the corresponding $B$--$H$ trajectory, which is further evaluated through the $B$--$H$ energy consistency and model compactness metrics.

\begin{table}[!t]
\centering
\caption{Overall sequence, energy, peak-field errors, energy-sign mismatch and parameter size compared with baseline models}
\label{tab:comparison_models}
\footnotesize
\setlength{\tabcolsep}{2.0pt}
\renewcommand{\arraystretch}{1.08}
\begingroup
\setlength{\aboverulesep}{0pt}
\setlength{\belowrulesep}{0pt}
\begin{tabular}{@{}lGGGGccc@{}}
\toprule
\multirow{2}{*}{\textbf{Model}}
& \multicolumn{2}{G}{\metricgroup[0.62in]{Sequence Error}}
& \multicolumn{2}{G}{\metricgroup[0.62in]{Energy Error}}
& \multirow{2}{*}{\begin{tabular}[c]{@{}c@{}}\textbf{Param.}\\\textbf{Size}\end{tabular}}
& \begin{tabular}[c]{@{}c@{}}\textbf{Peak}\\\textbf{Field}\\\textbf{Error}\end{tabular}
& \begin{tabular}[c]{@{}c@{}}\textbf{Energy}\\\textbf{Sign}\\\textbf{Mismatch}\end{tabular} \\
& \begin{tabular}[c]{@{}c@{}}Mean\\(\%)\end{tabular}
& \begin{tabular}[c]{@{}c@{}}95th\\(\%)\end{tabular}
& \begin{tabular}[c]{@{}c@{}}Mean\\(\%)\end{tabular}
& \begin{tabular}[c]{@{}c@{}}95th\\(\%)\end{tabular}
& & (\%) & (\%) \\

\midrule
\thistool & 13.40 & 31.39 & 1.92 & 7.60 & 4777 & 6.12 & 1.91 \\
MagLearn2 & 12.88 & 28.47 & 4.06 & 16.46 & 860929 & 5.67 & 8.39 \\
LSTM & 106.31 & 282.36 & 18.81 & 73.76 & 2399 & 49.21 & 38.15 \\
HARDCORE & 19.28 & 50.84 & 2.24 & 7.86 & 1629 & 10.35 & 4.33 \\
MMINN & 26.83 & 63.97 & 4.81 & 19.88 & 1084 & 17.50 & 6.22 \\
GRU & 18.44 & 46.17 & 3.68 & 16.62 & 4758 & 6.49 & 8.79 \\
Transformer & 364.92 & 930.32 & 23.43 & 87.68 & 2105 & 27.99 & 28.55 \\
\bottomrule
\end{tabular}
\endgroup
\end{table}

The advantage of \thistool\ is more evident when the trajectory-level metrics are considered together with sequence accuracy. Although its pointwise sequence error is close to that of MagLearn2, \thistool\ achieves the lowest mean and 95th-percentile energy errors, together with the lowest energy-sign mismatch rate among all evaluated models. This indicates that the proposed architecture does not merely fit the time-domain $H(t)$ waveform, but better preserves the accumulated energy behavior and traversal direction of the reconstructed $B$--$H$ trajectory. The result is consistent with the intended role of the energy-aware regularization and the hybrid local-global representation.

%The advantage of \thistool\ is more evident in energy-related metrics. It achieves the lowest mean and 95th-percentile energy errors, at 1.92\% and 7.60\%, respectively, whereas HARDCORE gives 2.24\% and 7.86\%, and MagLearn2 gives 4.06\% and 16.46\%. The energy-sign mismatch rate is also lower for \thistool\ than for all six models. These results indicate that the proposed physics-guided architecture and energy-aware training improve the consistency of the accumulated $B$--$H$ work, even when MagLearn2 gives slightly lower pointwise sequence errors.

Model compactness provides another important distinction. \thistool\ uses only 4777 trainable parameters, which is approximately $1/180$ of the parameter size of MagLearn2. The GRU has a comparable parameter size of 4758. However, its mean and 95th-percentile sequence errors increase to 18.44\% and 46.17\%, respectively, while its mean and 95th-percentile $B$--$H$ energy consistency errors increase to 3.68\% and 16.62\%. This comparison indicates that the performance of \thistool\ is not achieved simply through increased model capacity, but through the integration of local dynamic-state propagation, global hysteresis-context extraction, and $B$--$H$ energy consistency regularization. The remaining models exhibit larger errors, particularly in the 95th-percentile cases, suggesting that compact architectures without the combined dynamic, hysteresis-memory, and trajectory-consistency representations provide less favorable performance under the evaluated transient magnetization prediction conditions.

The behavior of HARDCORE highlights the importance of task-aligned physical structure. Although HARDCORE introduces an area-based intermediate $B$--$H$ representation for core-loss estimation \cite{ref_hardcore}, it does not explicitly model boundary-state initialization, hysteresis-memory propagation, or open-trajectory energy consistency. As a result, its performance is more material dependent: it can achieve competitive energy errors for some materials, but shows larger tail errors for others under the same evaluation setting. Such variability limits its robustness for engineering use, where a compact material model is expected to provide reliable transient prediction across different ferrites and operating conditions. In contrast, \thistool\ maintains a consistent balance of sequence accuracy, energy error, and energy-sign mismatch across materials.

\begin{figure*}[!b]
    \centering
    \includegraphics[width=\textwidth]{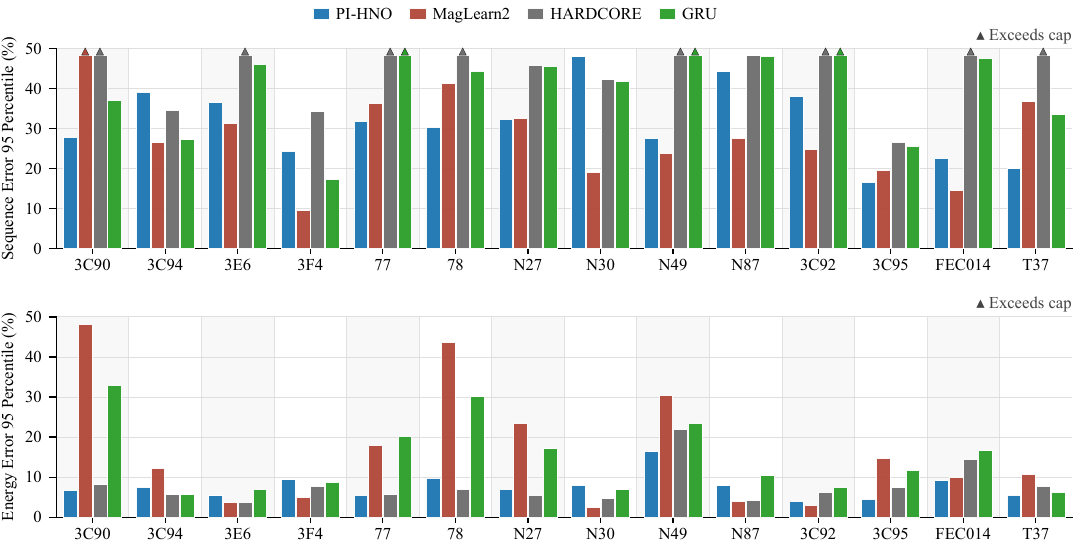}
    \caption{Material-wise 95th-percentile sequence (top) and energy (bottom) errors for \thistool, MagLearn2, HARDCORE, and GRU. Triangles mark values above the 50\% display limit.}
    \label{fig:material_comparison}
\end{figure*}

Fig.~\ref{fig:material_comparison} further compares the material-wise 95th-percentile sequence and energy errors for \thistool\ with those of the three strongest models. For the 95th-percentile sequence error, the comparison shows noticeable variation across materials. MagLearn2 gives lower 95th-percentile sequence errors for several materials, such as 3C94, 3F4, N30, and FEC014, whereas \thistool\ is lower for 3C90, 77, 78, and T37. The two models therefore show comparable but material-dependent sequence-prediction behavior, rather than one model uniformly dominating the other. In contrast, HARDCORE and GRU reach or exceed the 50\% display cap for multiple materials, indicating weaker control of the largest sequence errors under the same evaluation setting.

The material-wise energy comparison shows a clearer trend. MagLearn2 exhibits large 95th-percentile energy errors for several materials, including 3C90, 78, N27, and N49, with the 3C90 value being the highest and almost exceeding the display limit. \thistool\ also gives lower 95th-percentile energy errors than HARDCORE for 7 of the 14 materials. In contrast, the 95th-percentile energy error of \thistool\ remains below 10\% for all materials except N49. This material-wise behavior supports the aggregate results in Table~\ref{tab:comparison_models}: \thistool\ does not always minimize pointwise sequence error, but it provides the most consistent overall balance across the evaluated metrics.

\subsection{Ablation Study and Physical Interpretation}

Table~\ref{tab:ablation_results} reports the ablation results for the variants defined in Fig.~\ref{fig:ablation_design}. Each ablation increases the reported errors relative to \thistool, indicating that the tested components contribute complementary information to the transient $B$--$H$ prediction.

\begin{table}[!t]
\centering
\caption{Overall sequence, energy, peak-field errors, energy-sign mismatch and parameter size of the ablation study}
\label{tab:ablation_results}
\footnotesize
\setlength{\tabcolsep}{2.0pt}
\renewcommand{\arraystretch}{1.08}
\begingroup
\setlength{\aboverulesep}{0pt}
\setlength{\belowrulesep}{0pt}
\begin{tabular}{@{}lGGGGccc@{}}
\toprule
\multirow{2}{*}{\textbf{Variant}}
& \multicolumn{2}{G}{\metricgroup[0.62in]{Sequence Error}}
& \multicolumn{2}{G}{\metricgroup[0.62in]{Energy Error}}
& \multirow{2}{*}{\begin{tabular}[c]{@{}c@{}}\textbf{Param.}\\\textbf{Size}\end{tabular}}
& \begin{tabular}[c]{@{}c@{}}\textbf{Peak}\\\textbf{Field}\\\textbf{Error}\end{tabular}
& \begin{tabular}[c]{@{}c@{}}\textbf{Energy}\\\textbf{Sign}\\\textbf{Mismatch}\end{tabular} \\
& \begin{tabular}[c]{@{}c@{}}Mean\\(\%)\end{tabular}
& \begin{tabular}[c]{@{}c@{}}95th\\(\%)\end{tabular}
& \begin{tabular}[c]{@{}c@{}}Mean\\(\%)\end{tabular}
& \begin{tabular}[c]{@{}c@{}}95th\\(\%)\end{tabular}
& & (\%) & (\%) \\
\midrule
\thistool & 13.40 & 31.39 & 1.92 & 7.60 & 4777 & 6.12 & 1.91 \\
A1 & 17.77 & 42.19 & 2.86 & 9.61 & 2241 & 6.78 & 2.18 \\
A2 & 20.42 & 49.10 & 3.45 & 12.50 & 3817 & 17.80 & 5.57 \\
A3 & 17.53 & 40.09 & 3.73 & 12.66 & 4777 & 7.15 & 2.94 \\
A4 & 19.01 & 43.65 & 3.47 & 12.18 & 4665 & 7.71 & 2.77 \\
A5 & 24.83 & 75.68 & 3.26 & 12.49 & 4649 & 7.22 & 2.31 \\
\bottomrule
\end{tabular}
\par\vspace{0.6mm}
\endgroup
\end{table}

Removing the global branch (A1) reduces the parameter size from 4777 to 2241, but increases the mean sequence error from 13.40\% to 17.77\% and the 95th-percentile sequence error from 31.39\% to 42.19\%. The mean and 95th-percentile energy errors also increase from 1.92\% and 7.60\% to 2.86\% and 9.61\%, respectively. The degradation suggests that the complete $B(t)$ trajectory provides additional magnetic-history information relevant to the magnetization process. Physically, this is consistent with the path-dependent nature of quasi-static hysteresis: reversal locations, flux-density extrema, dc offset, and minor-loop context cannot be fully recovered from a short local history alone.

Removing the local recurrent branch (A2) produces the largest peak-field error and energy-sign mismatch rate among the ablations, at 17.80\% and 5.57\%, respectively. It also increases the 95th-percentile sequence and energy errors to 49.10\% and 12.50\%. This indicates that the local branch is essential for initializing the boundary magnetic state and preserving the rate-dependent eddy-current behavior over the subsequent excitation interval. Without this local dynamic state, the model loses accuracy in the field amplitude and is more likely to predict the wrong direction of accumulated magnetization.

Removing the incremental $\Delta B$ input (A3) gives the lowest mean sequence error among the ablated variants. Still, it produces the largest mean and 95th-percentile $B$--$H$ energy consistency errors, at 3.73\% and 12.66\%, respectively. This behavior is physically meaningful. The instantaneous value of $B(t)$ identifies the operating point on the flux-density axis, whereas $\Delta B$ provides information on the direction of local variation and the excitation rate. Removing $\Delta B$ therefore reduces the model's ability to capture transient magnetic effects associated with changing flux-density excitation. As a result, the predicted waveform can still achieve relatively small pointwise errors, while the accumulated $B$--$H$ work along the reconstructed trajectory becomes less consistent.

Removing the future GRU (A4) raises the mean sequence error to 19.01\% and the 95th-percentile energy error to 12.18\%. The increased errors indicate that $H(t)$ prediction requires sequential magnetic-state propagation along the input $B(t)$ trajectory, rather than a direct $B(t)$-to-$H(t)$ mapping. This is consistent with dynamic hysteresis, in which the instantaneous field depends on both the current flux density and the previously accumulated hysteretic state.

Finally, bypassing temperature- and start-position-conditioned hidden-state initialization (A5) gives the largest sequence degradation, with mean and 95th-percentile sequence errors of 24.83\% and 75.68\%, respectively. Since temperature and start-position information are still provided directly to the future GRU in this variant, the result does not imply that these inputs are unimportant. Rather, it shows that their placement is important. In this architecture, temperature-dependent material behavior and prediction-boundary context are more effectively introduced via the initial hidden state than as stepwise future inputs.

Overall, the ablation results support the intended division of roles in \thistool. The global attention branch supplies a quasi-static hysteresis context. The local recurrent branch propagates the rate-dependent dynamic response. The $\Delta B$ input provides local excitation-change information. The future GRU maintains the evolving prediction state, and the conditioning projections initialize the model with temperature and boundary-position context. The full model provides the best overall balance of sequence accuracy, accumulated-energy consistency, peak-field prediction, directionality of energy dissipation, and parameter size.

\section{Conclusion}\label{sec:conclusion}
This work proposed \thistool, a compact physics-informed material-specific neural model for core-loss-oriented transient magnetization prediction. Given the complete $B(t)$ excitation trajectory, the recent $H(t)$ history before the prediction interval, temperature, and operating-condition information, \thistool\ predicts the subsequent $H(t)$ response and the corresponding reconstructed $B$--$H$ trajectory. The proposed framework incorporates magnetic knowledge through a physics-guided decomposition of transient magnetic response, a Preisach-inspired global representation for capturing hysteresis-related waveform context, and a $B$--$H$ energy consistency regularization term during training. Specifically, the local recurrent branch learns a boundary-conditioned representation of rate-dependent magnetic response, while the global branch exploits the memory aggregation principle of Preisach hysteresis to extract long-range excitation dependencies from the complete $B(t)$ trajectory.

Evaluation using material-specific models for 14 ferrite materials from the MagNetX transient dataset demonstrated that \thistool\ achieved mean sequence and energy errors of 13.40\% and 1.92\%, respectively, with corresponding 95th-percentile errors of 31.39\% and 7.60\%. These results were obtained with only 4777 trainable parameters per material-specific model, confirming the compactness of the proposed architecture for transient magnetization modeling. Comparison with the other six magnetic modeling approaches indicates its success in balancing sequence accuracy, accumulated-energy consistency, peak-field prediction, directionality of energy dissipation, and parameter size. The ablation study further demonstrated that the proposed architectural components and conditioning strategies provide complementary contributions to prediction accuracy and accumulated-energy consistency, supporting the effectiveness of the physics-guided design.

Future work will focus on embedding \thistool\ into transient electromagnetic and circuit-field simulation workflows and validating its impact on device-level loss prediction, thermal assessment, and high-frequency magnetic component optimization across a broader range of excitation and operating conditions.

%%*************************************************************************

\bibliographystyle{IEEEtran}
\bibliography{reference}

\end{document}